# Orthogonal Matrices for MBAT Vector Symbolic Architectures, and a "Soft" VSA Representation for JSON

**Stephen I. Gallant**
sgallant@textician.com

**Textician**

Cambridge, MA 02138
February 7, 2022

**Abstract**

Vector Symbolic Architectures (VSAs) give a way to represent a complex object as a single fixed-length vector, so that similar objects have similar vector representations. These vector representations then become easy to use for machine learning or nearest-neighbor search. We review a previously proposed VSA method, MBAT (Matrix Binding of Additive Terms), which uses multiplication by random matrices for binding related terms. However, multiplying by such matrices introduces instabilities which can harm performance. Making the random matrices be orthogonal matrices provably fixes this problem. With respect to larger scale applications, we see how to apply MBAT vector representations for any data expressed in JSON. JSON is used in numerous programming languages to express complex data, but its native format appears highly unsuited for machine learning. Expressing JSON as a fixed-length vector makes it readily usable for machine learning and nearest-neighbor search. Creating such JSON vectors also shows that a VSA needs to employ binding operations that are non-commutative. VSAs are now ready to try with full-scale practical applications, including healthcare, pharmaceuticals, and genomics.

## 1 Introduction

We are interested in Vector Symbolic Architectures (VSAs)[1] which represent a complex object as a single fixed-length vector so that similar objects have similar vector representations. VSAs use *distributed representations* that encode information over many or all vector dimensions, rather than localist representations that encode information over a few special words or vector positions. Arbitrarily complex objects, such as a sentence and its parse, can be represented as a single distributed vector. VSAs give a way to represent similar objects, for example "The large elephant" vs. "The huge elephant", as similar vectors, when measured by vector distance.

By way of notation, we assume a fixed $n$-dimensional space, where a typical value for $n$ might be 300 dimensions or 1,000 dimensions. Vectors, **V**, are column vectors of dimension $n$, with subscripts identifying particular vectors, such as $\mathbf{V}_A$ or $\mathbf{V}_B$. Matrices, **M**, are $n$ by $n$, again with subscripts used to identify individual matrices.

---

[1] Also called Hyperdimensional Computing, HDC.

Thanks to Phil Culliton and Dmitri Rachkovskij for helpful suggestions.

For notational convenience, we incorporate a transpose into the dot or inner product:

$V_A \bullet V_B$ represents $V_A^{Transpose} \bullet V_B$.

An important property of high dimensional vectors is that vector sums "remember" which individual vectors were added. For example, if we add 20 random vectors to form a vector sum $V$, then we can test whether any vector, $V_A$ (or a vector similar to $V_A$) was among the 20 vectors: we merely compute $V_A \bullet V$. A high positive value indicates that $V_A$ was in the sum. This *Vector Sum Memory* property is counter-intuitive, because it doesn't hold for scalars: if we add 20 numbers together and get 117, there is no way to tell whether one of those numbers was 14. The proof of Vector Sum Memory is straightforward:

$$V_A \bullet V = \begin{cases} \|V_A\|^2 + \text{noise} & \text{if } V_A \text{ is in the sum for } V \\ 0 + \text{noise} & \text{otherwise.} \end{cases}$$

As dimensions increase, the signal dominates the noise. See [Gallant & Okaywe (2013)] for further analysis and simulations. To take an example (from Table 2 in that paper), suppose we sum 100 random vectors having entries ±1 to get vector $V$. If we now take 100,000 other random vectors, the probability that *all* 100,000 have lower dot products with $V$ than any of the original 100 vectors is 98% if we are working in at least 7,000 dimensions.

Previously we proposed a method for constructing VSAs called Matrix Binding of Additive Terms (MBAT) [Gallant & Okaywe (2013)]. The basic idea is to associate (bind) sets of objects by multiplying the sum of their corresponding vectors by a binding matrix. For each type of binding, the binding matrix can be a fixed random matrix. For example, if $M_{actor}$ is the binding matrix (possibly randomly generated) associated with *Actor* in a sentence parse, and $V_{word}$ is the vector for a given word, then we could generate the vector for Actor = "The clever researcher" by

$$V = M_{actor} ( V_{The} + V_{clever} + V_{researcher} )$$
$$= M_{actor} V_{The} + M_{actor} V_{clever} + M_{actor} V_{researcher}.$$

We can recursively apply this scheme to represent arbitrarily complex nested concepts. Moreover, we can use Vector Sum Memory to determine whether, for example, $M_{actor} V_{researcher}$ is likely to be a part of $V$ by taking the dot product of the two vectors and noting whether it is large.

## 2   Binding and Complex Structures

Binding is useful for representing structure. We need it to differentiate "The clever researcher saw a large elephant" from "The clever elephant saw a large researcher." Both sentences use the same words, so binding is needed to specify whether *clever* refers to *researcher* or to *elephant*.

The motivation for MBAT was that the human brain has many recurrent connections; therefore neuron influences are better modeled by a matrix with many recurrent connections --- not a triangular matrix with only feedforward influences. This suggested using matrix multiplication for binding, which works. See [Gallant & Okaywe (2013)] for more details, including discussion of representing sequences and capacity simulations.

This paper has two goals:

- The first goal is to repair an instability problem with MBAT representations, due to choosing random matrices for binding matrices.

- The second goal is to give a practical example of using MBAT to represent highly-structured data descriptions. For this, we will show how to employ MBAT to convert JSON data descriptions to fixed-length vectors, which allows JSON to be easily used for machine learning and similarity search. JSON is frequently used in computer science for specifying and representing things like data inputs to software applications. Normally there is no notion of closeness between two JSON descriptions.

## 3  A problem with MBAT

MBAT representations can handle arbitrarily complex data relationships by binding additive terms, where terms may themselves be the results of other bindings. For example,

$$\mathbf{V} = \mathbf{M}_{sentence} ( \mathbf{M}_{actor} ( \mathbf{V}_{The} + \mathbf{V}_{clever} + \mathbf{V}_{researcher} ) + \mathbf{M}_{verb} (\mathbf{V}_{saw}) + \mathbf{M}_{object} ( \mathbf{V}_{The} + \mathbf{V}_{large} + \mathbf{V}_{elephant})).$$

The problem is that when using random binding matrices, entries of $\mathbf{M}_{actor}$ might be much larger than entries of $\mathbf{M}_{object}$ so that the object group is ignored. Such instability becomes more problematic when multiplying random matrices, such as in $\mathbf{M}_{sentence} \mathbf{M}_{actor}$ or in more deeply nested constructs that result in many matrix multiplications. This problem can be particularly acute when representing ordered sequences by repeated matrix multiplications. For example the vector for the sequence [A, B, C] becomes:

$$\mathbf{V}_{sequence} = \mathbf{M}_{SEQ} \mathbf{V}_A + \mathbf{M}_{SEQ} \mathbf{M}_{SEQ} \mathbf{V}_B + \mathbf{M}_{SEQ} \mathbf{M}_{SEQ} \mathbf{M}_{SEQ} \mathbf{V}_C.$$

One possible fix for this is to normalize vectors after each binding operation, but there is a better way: use random orthogonal binding matrices.

## 4  Random orthogonal matrices as binding matrices

An *orthogonal matrix* is a matrix whose columns, $\mathbf{M}(*,i)$, are mutually orthogonal and have length 1. Using orthogonal binding matrices solves the stability problem.

We first review important properties of orthogonal matrices, which are defined by

$$\mathbf{M}(*,i) \bullet \mathbf{M}(*,j) = \begin{cases} 0 \ for \ i \neq j, \ and \\ 1 \ for \ i = j. \end{cases}$$

A. *Orthogonal matrices preserve vector length*: $\| \mathbf{MV} \| = \| \mathbf{V} \|$. Thus multiplication by an orthogonal matrix just gives a rotation in its high-dimensional space. This property is well-known, but for completeness and to help with intuition we give the simple proof. Note that multiplying a matrix with a vector is the same as summing the columns of the matrix, with each column weighted by the corresponding vector component.[2]

$$\|\mathbf{MV}\|^2 = \mathbf{MV} \bullet \mathbf{MV}$$

$$= ( \sum_i \mathbf{M}(*,i) \mathbf{V}_i ) \bullet ( \sum_j \mathbf{M}(*,j) \mathbf{V}_j )$$

$$= \sum_{i \neq j} \mathbf{V}_i \mathbf{V}_j ( \mathbf{M}(*,i) \bullet \mathbf{M}(*,j) ) + \sum_i \mathbf{V}_i \mathbf{V}_i ( \mathbf{M}(*,i) \bullet \mathbf{M}(*,i) ) \quad \text{by rearranging}$$

$$= 0 + \sum_i \mathbf{V}_i \mathbf{V}_i (1) \quad \text{by orthogonality}$$

$$= \mathbf{V} \bullet \mathbf{V} = \|\mathbf{V}\|^2$$

---

[2] This is not how we were taught to do matrix multiplication in school.

Because magnitudes are non-negative, having equal squares proves $\|\mathbf{MV}\| = \|\mathbf{V}\|$. This property, along with the associativity, in general, of matrix multiplication[3], makes binding by many orthogonal matrices stable:

$$\|\mathbf{M_A M_B M_C V}\| = \|\mathbf{M_A (M_B (M_C V))}\| = \|\mathbf{V}\|.$$

B. *For orthogonal matrices, their inverse is just their transpose*: $\mathbf{M^{-1}} = \mathbf{M^{transpose}}$. This follows from $\mathbf{M^{transpose}\ M} = \mathbf{I}$.

Inverses can come in handy when computing which components are bound by a particular binding matrix. For example, if we want to know whether $\mathbf{MV_A}$ or $\mathbf{MV_B}$ is in resulting vector $\mathbf{V}$, we can compute $\mathbf{M^{-1}\ V}$ and dot it with $\mathbf{V_A}$ and $\mathbf{V_B}$, rather than having to compute both $\mathbf{MV_A}$ and $\mathbf{MV_B}$.

Tissera and McDonnell [2014] previously proposed using orthogonal matrices with a variation on MBAT, solely because of orthogonal matrices having readily available inverses.

C. *It's easy to generate random orthogonal matrices*[4]. For each column *j*, in order:

    a. Start with a random vector, $\mathbf{V}_j$.

    b. For each previous column, $\mathbf{V}_k$, use Gram-Schmidt elimination, replacing $\mathbf{V}_j$ by $\mathbf{V}_j - (\mathbf{V}_j \bullet \mathbf{V}_k)\ \mathbf{V}_k$**,** to make sure $\mathbf{V}_j$ is orthogonal to $\mathbf{V}_k$.

    (We can also see by induction that the revised $\mathbf{V}_j$ remains orthogonal to all $\mathbf{V}_p$ for p<k, because we are subtracting from $\mathbf{V}_j$ a multiple of $\mathbf{V}_k$ which, by induction, is orthogonal to $\mathbf{V}_p$.)

    c. Normalize $\mathbf{V}_j$. (In the exceedingly unlikely case that $\mathbf{V}_j$ is the 0 vector, go back to step a.)

(It can be shown that an orthogonal matrix must also have orthonormal rows, but we do not make use this property.)

By property A, using length-preserving orthogonal matrices preserves the lengths of bound terms in MBAT representations, regardless of the complexity of the structure that is encoded. This keeps complex structures from exploding in magnitude and solves our problem.

## 5   Application: representing JSON data by MBAT VSAs

A tough test for representing complex objects is representing a JSON data description as a fixed-length vector using MBAT.

JSON (www.json.org) is a language independent way to express data with arbitrarily complex organization. It uses objects, arrays, and values where:

- An *object* is an unordered set of name/value pairs.
- An *array* is an ordered collection of values.
- A *value* can be a *string* in double quotes, or a *number*, or true or false or null, or an *object* or an *array*. These structures can be nested.

---

[3] Matrix multiplication is associative, but not commutative.
[4] Or just use Python ortho_group.rvs.

For example (shortened from www.json.org):

```
{
    "squadName": "Super hero squad",
    "homeTown": "Metro City",
    "active": true,
    "members": [
        {
          "name": "Molecule Man",
          "age": 29,
          "secretIdentity": "Dan Jukes",
          "powers": [
                  "Radiation resistance",
                  "Turning tiny",
                  "Radiation blast"
          ]
        },
        {
          "name": "Madame Uppercut",
          "age": 39,
          "secretIdentity": "Jane Wilson",
          "powers": [
                  "Million tonne punch",
                  "Damage resistance",
                  "Superhuman reflexes"
          ]
        }
    ]
}
```

Using native JSON objects directly for machine learning looks quite daunting, to say the least. Similarly, if we only want to find the closest matches from a collection of JSON descriptions, we could hack together some ad hoc search program but, again, this would be unappealing.

However, using MBAT representations we can convert any JSON description to a fixed-length vector that is "soft" (preserves similarities), and then do machine learning or nearest-neighbor searches. The conversion is simple:

- Represent a JSON *object* with MBAT by converting a name/value pair to a binding matrix times the corresponding value vector: $\mathbf{M}_{name} \, ( \, \mathbf{V}_{value} \, )$.

- Represent a JSON *array* as a vector representing a sequence, as described in Section 3 above. The resulting sequence vector may optionally be normalized.

- Represent a JSON *value* by a vector where:

    o *Strings* are represented by the sum of vectors for words in the string:

    $$\mathbf{V}_{\text{"the smart researcher"}} \;=\; \mathbf{V}_{\text{"the"}} + \mathbf{V}_{\text{"smart"}} + \mathbf{V}_{\text{"researcher"}}.$$

    To represent a vector for a word, we can use a normalized random vector or a vector from some precomputed dictionary such as Word2Vec [Mikolov (2013)] or Glove vectors [Pennington (2014)]. If order counts in the string, then we can use a sequence vector. Optionally, we can normalize vectors for strings.

- *Numbers* are represented by a normalized random vector when there is no similarity between different numbers, such as ID numbers. If we want numerical similarity, we can use a set of thresholds, sometimes called a *thermometer code*:

  $\mathbf{V}_{Sum}$ = normalized random vector $\mathbf{V}_{number}$

  $+ \mathbf{V}_{very\_low}$ if quantity ≥ very low threshold

  $+ \mathbf{V}_{low}$ if quantity ≥ low threshold

  $+ \ldots$

  $+ \mathbf{V}_{high}$ if quantity ≥ high threshold

  Each of the individual threshold vectors is a normalized random vector, and we also normalize the resulting sum.

  - For each of **true**, **false** and **null**, use fixed, normalized random vectors.

Thus we can convert any JSON description into an MBAT fixed-length vector. For example:

$\mathbf{V} = \mathbf{M}_{SquadName} (\mathbf{V}_{Super} + \mathbf{V}_{hero} + \mathbf{V}_{squad}) + \mathbf{M}_{HomeTown} (\mathbf{V}_{Mexico} + \mathbf{V}_{City}) + \mathbf{M}_{Active} (\mathbf{V}_{true})$

$+ \mathbf{M}_{members} ($

$\quad \mathbf{M}_{SEQ} ($

$\quad\quad \mathbf{M}_{name} (\mathbf{V}_{Molecule} + \mathbf{V}_{Man}) +$

$\quad\quad \mathbf{M}_{age} (\text{normalize}(\mathbf{V}_{number} + \mathbf{V}_{older\_than\_15} + \mathbf{V}_{older\_than\_25})) +$

$\quad\quad \mathbf{M}_{Secret\_identity} (\mathbf{V}_{Dan} + \mathbf{V}_{Jukes}) +$

$\quad\quad \mathbf{M}_{powers} ($

$\quad\quad\quad \mathbf{M}_{SEQ} (\mathbf{V}_{Radiation} + \mathbf{V}_{resistance})$

$\quad\quad\quad + \mathbf{M}_{SEQ} \mathbf{M}_{SEQ} (\mathbf{V}_{Turning} + \mathbf{V}_{tiny})$

$\quad\quad\quad + \mathbf{M}_{SEQ} \mathbf{M}_{SEQ} \mathbf{M}_{SEQ} (\mathbf{V}_{Radiation} + \mathbf{V}_{blast})$

$\quad\quad )$

$\quad )$

$\quad + \mathbf{M}_{SEQ} \mathbf{M}_{SEQ} ($

$\quad\quad \mathbf{M}_{name} (\mathbf{V}_{Madame} + \mathbf{V}_{Uppercut}) +$

… etc.

$)$

Notice that some of the 29 added vectors used in computing the resulting vector may be complex, such as $\mathbf{M}_{members} \mathbf{M}_{SEQ} \mathbf{M}_{powers} \mathbf{M}_{SEQ} \mathbf{M}_{SEQ} \mathbf{M}_{SEQ} \mathbf{V}_{Radiation}$, but it has the same length as just $\mathbf{V}_{\text{Radiation}}$, and is simply a rotated version of $\mathbf{V}_{\text{Radiation}}$. Note, also, that this term is differentiated from the term we get by shifting one of the $\mathbf{M}_{\text{SEQ}}$ binding operators: $\mathbf{M}_{\text{members}}$ $\mathbf{M}_{\text{SEQ}}$ $\mathbf{M}_{\text{SEQ}}$ $\mathbf{M}_{\text{powers}}$ $\mathbf{M}_{\text{SEQ}}$ $\mathbf{M}_{\text{SEQ}}$ $\mathbf{V}_{\text{Radiation}}$, which would place "radiation" in the second power of Madame Uppercut rather than the third power of Molecule Man. However, if binding were commutative, both terms would be equal. This illustrates the necessity for complex objects that *binding be non-commutative*. Plate [2003] (Chapter 3.6.7) was aware of this issue and proposed several non-commutative alternative formulations to his HRR system, including matrix multiplication.

Presumably other VSA techniques having non-commutative binding operators can also represent JSON descriptions, but it must be verified that deeply embedded terms don't cause problems.

# 6   Prior research

Vector Symbolic Architectures are an expanding field, with many theoretical and applied directions. Some classic papers include Kanerva [1988, 2009], Plate [2003], Gayler [1998], Levy and Gayler [2008], and Rachkovskij [2001]. Luckily, there is an excellent two-part review by Kleyko, Rachkovskij, and associates [2021a,b] that is thorough, recent, and readily available. We refer the reader to these two papers for a better job presenting prior research than we are able to give.

# 7   Discussion and applications

MBAT in particular, and VSAs in general, can be viewed as similarity-preserving hash codes for complex structures, where resulting hashes are amenable to machine learning and to searching for nearest neighbors.

With respect to neural information processing, the Vector Sum Memory principle gives a way to construct a reliable memory using many independent and unreliable components, such as neurons. We would be astounded if something analogous were not used by brains.

The JSON example shows that we can represent complex data structures, and that in order to represent them, we need to use VSA binding operations that are non-commutative.

The ability to represent complex objects by a distributed fixed-length vector suggests many applications for MBAT, including:

- *Healthcare:* We can represent important parts of the Electronic Health Record (EHR) as a fixed-length vector. This would permit easy modeling of who is at risk for a particular condition, plus many other modeling tasks.

- *Pharmaceutical Development:* Having fixed-length vector representations for individual EHRs would permit, for a particular treatment, modeling who would benefit and who would experience adverse events.

- *Genomics:* We should be able to represent genomic sequences as fixed-length vectors for modeling, where the binding mechanism can specify ranges for sequence annotations. This would permit predictive modeling of which genome sequences are associated with particular conditions or diseases.

In conclusion, VSAs appear ready to try with full-scale, practical applications.